\documentclass[lettersize,journal]{IEEEtran}
\usepackage{amsmath,amsfonts}
\usepackage{algorithmic}
\usepackage{algorithm}
\usepackage{array}
\usepackage{xcolor}
\usepackage{colortbl}
\usepackage{textcomp}
\usepackage{stfloats}
\usepackage{url}
\usepackage{verbatim}
\usepackage{graphicx,float}
\usepackage{authblk}
\usepackage{cite}
\usepackage{caption}
\usepackage{subcaption}
\usepackage{multicol}
\usepackage{multirow}
\usepackage{rotating} 
\usepackage{booktabs}  
\usepackage[hidelinks]{hyperref}
\usepackage{threeparttable}
\usepackage{tikz}
\usetikzlibrary{shapes.geometric}
\usetikzlibrary{positioning}
\begin{document}

\title{Analysis of the Unscented Transform for Cooperative Localization with only Inter-vehicle Ranging Information }

\author[1]{Uthman Olawoye}
\author[2]{Cagri Kilic}
\author[1]{Jason N. Gross}
\affil[1]{Department of Mechanical and Aerospace Engineering\\
West Virginia University\\
Morgantown, WV}
\affil[2]{Department of  Aerospace Engineering\\
Embry-Riddle Aeronautical University\\
Daytona Beach, FL}



\maketitle

\begin{abstract}
Cooperative localization in multi-agent robotic systems is challenging, especially when agents rely on limited information, such as only peer-to-peer range measurements. Two key challenges arise: utilizing this limited information to improve position estimation; handling uncertainties from sensor noise, nonlinearity, and unknown correlations between agents’ measurements; and avoiding information reuse. This paper examines the use of the Unscented Transform (UT) for state estimation for a case in which range measurement between agents and covariance intersection (CI) is used to handle unknown correlations. Unlike Kalman Filter approaches, CI methods fuse complete state and covariance estimates. This makes formulating a CI approach with ranging-only measurements a challenge. To overcome this, UT is used to handle uncertainties and formulate a cooperative state update using range measurements and current cooperative state estimates. This introduces information reuse in the measurement update. Therefore, this work aims to evaluate the limitations and utility of this formulation when faced with various levels of state measurement uncertainty and errors.
\end{abstract}


\section{Introduction}
Cooperative localization has emerged as a viable strategy for increasing the accuracy and resilience of multi-robot systems' localization \cite{avinashmultirobot}. Cooperative localization provides a more reliable estimation of each robot's position by combining relative measurements across robots and merging sensor data from different sources \cite{gao2019cooperative}. Yet, coordination and collaboration can become difficult given the increasing processing and communication costs in large groups of multi-robot and swarm robotic systems \cite{bailey2011decentralised}. Several mobile robot tasks, such as exploration, navigation, object identification and tracking, and map construction, require an accurate localization estimate. By leveraging relative measurements amongst robots and exchanging their state estimations and covariances \cite{luft2018recursive}, cooperative localization improves localization performance significantly when using minimal sensors onboard \cite{gutierrez2022pseudo}. Cooperative localization in multi-robot systems can be divided into centralized and decentralized techniques. 

Decentralized techniques, in particular, have become more common because of their lower computing and communication costs and their resistance to failure, though they tend to be less optimal in some aspects.\cite{carrillo2013decentralized,lassoued2014mobile}. For example, individual state estimations are often produced by combining proprioceptive and exteroceptive sensor readings. Nevertheless, GNSS coverage is restricted in urban and forested regions, resulting in poor localization performance in these conditions~\cite{merry2019smartphone}. In this respect, if some robots in the system are in an area where the GNSS signal is disrupted, others having GNSS access can exchange this information during relative updates~\cite{qu2011cooperative}. The decentralized architecture of the filter estimator also allows the robots to decouple their individual states to reduce computational costs. Moreover, the robots are equipped with UWB sensors that allow them to perform relative ranging measurement updates with other robots within a specified proximity~\cite{gutierrez2022pseudo}. This is done by coupling the states and covariances of the robots participating in the update.

When robots estimate poses independently, correlations between them are often lost, leading to inconsistent results in traditional methods like the Kalman filter. One solution is to avoid reusing information multiple times. Some methods have robots compute local estimates from their sensors and fuse them with other agents' data. Only the local estimate is shared to minimize correlation issues \cite{karam2006localization}. Another approach \cite{julier2017general} uses the CI Filter, which assumes full dependence between measurements and state vectors, acting as a conservative version of the Kalman filter. The Split Covariance Intersection Filter (SCIF) refines this by managing dependency configurations for more accurate estimations \cite{li2013split,wanasinghe2014decentralized}. Other methods use pose measurements via LiDAR \cite{fang2022lidar, hery2021consistent, olawoye2023uav} or cameras \cite{zhu2021distributed}, but these require extensive data processing. Low-cost Ultra-Wideband (UWB) sensors can provide relative ranging with centimeter precision \cite{zwirello2012uwb}. Range-only data, while useful, leads to non-linear functions that are difficult for fusion.
Ranging-only cooperative localization has been widely studied \cite{pierre2018range,burchett2019unscented,mcinerney2017cooperative,shi2019range,liu2018multi,araki2019range}, often relying on multiple UWB sensors for inter-agent distance measurements and techniques like triangulation. Fixed anchors are usually required to act as reference points, but dynamic agents can also serve as temporary anchors. However, reliance on range-only data introduces cumulative errors over time, and the Kalman Filter struggles with its nonlinearity and unknown sensor correlations. Traditional approaches requiring UWB-ranging infrastructure face limitations in dynamic environments where static anchors may not be feasible. These issues are compounded by potential information reuse, which distorts accuracy.
Our approach leverages the UT for state estimation and CI for data fusion to enable cooperative localization in ranging-only scenarios. The primary objective of this work is to rigorously assess the applicability and limitations of the UT in conjunction with CI for cooperative localization, particularly in situations where agents rely solely on range measurements between robots. By employing UT, we aim to manage uncertainties effectively and perform cooperative state updates that leverage range data and existing cooperative state estimates. 

The rest of this paper is organized as follows. Section~\ref{problem} defines the problem and introduces the notations, state representations, and mathematical equations. Section~\ref{method} details and explains the components of the methodology. Section~\ref{experimental_results} provides the initial test setup. Finally, Section~\ref {results} and \ref{conclusion} present the expected results, contributions, and insights for future work to improve the system.       

\section{Problem Statement}
\label{problem}
In this paper, we consider a multivehicle localization scenario inspired by \cite{akhihiero2024cooperative} involving two robotic agents, denoted as $\mathbf{A_1}$ and $\mathbf{A_2}$, operating in a 2D environment. Let the state vectors be defined as the state of $\mathbf{A_1}$ and $\mathbf{A_2}$ at time step \(k\) and represented as:
\begin{equation}
    \mathbf{x}_1^k = \begin{bmatrix} x_1^k & y_1^k & v_{x_1}^k &v_{y_1}^k \end{bmatrix}^\top,  \mathbf{x}_2^k = \begin{bmatrix} x_2^k & y_2^k & v_{x_2}^k &v_{y_2}^k \end{bmatrix}^\top
    \label{e1}
\end{equation}
where \( x^k \), \( y^k \),\(v_{x}^k\) and \(v_{y}^k\) denote the position and velocity of the respective vehicles at time step \( k \). where the numbered subscripts denote the robots. The robots are assumed to be point masses; thus, only the position is estimated, i.e., no orientation. The corresponding covariance matrices of their states are given by \( \mathbf{P}_1^k \) and \( \mathbf{P}_2^k \). The inter-vehicle range measurement at time step \( k \) is:
\begin{equation}
    r^k = \|\mathbf{x}_1^k - \mathbf{x}_2^k\| + u
\end{equation}
where \( u \sim \mathcal{N}(0, R) \) represents zero-mean Gaussian noise with variance \( R \).\\
The objective is to estimate the state of $\mathbf{A_2}$, \( \mathbf{x}_2^k \) and minimize the localization uncertainty, given:
\begin{itemize}
    \item Reliable Positioning updates for  \(\mathbf{A_1}\): 
   \begin{equation}
        \mathbf{y}_1^k = \mathbf{x}_1^k + \mathbf{w}_1, \quad \mathbf{w}_1 \sim \mathcal{N}(0, \mathbf{Q}_1)
   \end{equation} where \( \mathbf{y}_1^k \) is the positioning measurement and \( \mathbf{Q}_1 \) is the covariance of the position solution uncertainty.
    \item Sporadic and noisy Positioning updates for \(\mathbf{A_2}\):
    \vspace{-8pt}
    \begin{equation}
         \mathbf{y}_2^k = \mathbf{x}_2^k + \mathbf{w}_2, \quad \mathbf{w}_2 \sim \mathcal{N}(0, \mathbf{Q}_2)
    \end{equation} where \( \mathbf{Q}_2 \) is the covariance of $\mathbf{A_2}$’s position solution uncertainty.
    \item Inter-vehicle range measurements, \( r^k \), which are nonlinear functions of the vehicle states.
\end{itemize}

\begin{figure}[htb!]
    \centering
    \begin{tikzpicture}[scale=1.2]
    \draw[gray!30, thin] (-1,0) grid (6,4);

    \draw[blue, thick, -stealth] 
        (0.5,2) .. controls (2,2) and (4,1) .. (5.5,3.5)
        coordinate[pos=0.15] (r1k1)
        coordinate[pos=0.42] (r1k2)
        coordinate[pos=0.78] (r1k3)
        coordinate[pos=1.00] (r1k4);
    
    \draw[red, thick, dashed, -stealth] 
        (0.5,0.3) .. controls (2,1.7) and (4,0.7) .. (6,0.5)
        coordinate[pos=0.25] (r2k1)
        coordinate[pos=0.50] (r2k2)
        coordinate[pos=0.65] (r2k3)
        coordinate[pos=1.00] (r2k4);
    
    \draw[black, thick, densely dotted] (r1k1) -- (r2k1) node[midway, above, sloped] {$r_{1}$};
    \draw[black, thick, densely dotted] (r1k2) -- (r2k2) node[midway, above, sloped] {$r_{2}$};
    \draw[black, thick, densely dotted] (r1k3) -- (r2k3) node[midway, above, sloped] {$r_{3}$};
    \draw[black, thick, densely dotted] (r1k4) -- (r2k4) node[midway, above, sloped] {$r_{4}$};
    
    \foreach \x/\y/\label in {r1k1/blue/1, r1k2/blue/2, r1k3/blue/3, r1k4/blue/4} {
        \fill[black] (\x) circle (0.05);
    }
    \foreach \x/\y/\label in {r2k1/red/1, r2k2/red/2, r2k3/red/3, r2k4/red/4} {
        \fill[red] (\x) circle (0.05);
    }
    
    \fill[blue] (0.5,2) circle (0.1);
    \fill[red] (0.5,0.3) circle (0.1);
    \fill[blue] (5.5,3.5) circle (0.1);
    \fill[red] (6,0.5) circle (0.1);

    \node[blue] at (0,2) {$\mathbf{A_1}$};
    \node[red] at (0,0.3) {$\mathbf{A_2}$};
    
    \foreach \i in {1, 2, 3, 4} {
        \draw[blue, thick] (r1k\i) circle (0.2);  
    }
    
    \foreach \i/\p in {1/(0.3 and 0.1), 2/(0.4 and 0.15), 3/(0.5 and 0.2), 4/(0.8 and 0.25)} {
        \draw[red, thick] (r2k\i) ellipse \p;  
    }

    \end{tikzpicture}
    \caption{Two-Robot Localization Scenario with Inter-Vehicle Range Measurements}
    \label{fig:enter-lal}
\end{figure}
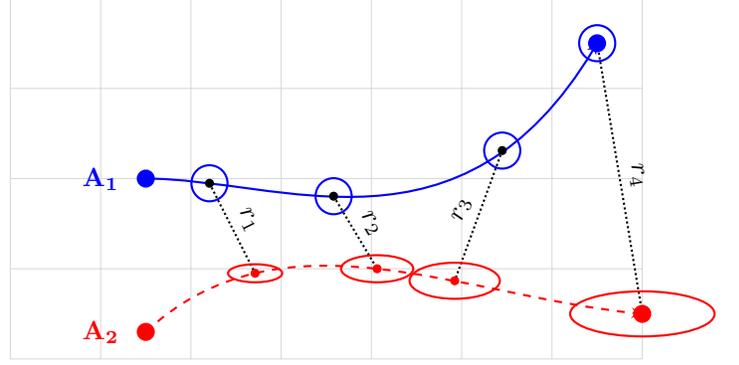

\section{Methodology}
\label{method}
The main objective, as defined above, is to improve the accuracy of state estimation using limited communication data. Specifically, range measurements and state estimates are shared between agents. 
The proposed method is evaluated using a 2D simulation of a toy example involving two robots: Robot $\mathbf{A_1}$ gets reliable and frequent absolute positioning updates, and Robot $\mathbf{A_2}$, whose state is more uncertain due to sporadic and noisy positioning updates. In our testing scenarios, the robots will move along predefined paths with $\mathbf{A_1}$ state being estimated with a simple Kalman Filter.
\subsection{Kalman Filter}
The Kalman Filter \cite{kalman1960new} is a widely-used recursive algorithm used for estimating the state of a dynamic system from a series of noisy measurements. The Kalman filter operates in two steps: prediction and update. The filter uses a system model to estimate the current state and uncertainty in the prediction step and incorporates new measurements to refine the state estimate in the update step. This balances out the uncertainty in the model and the measurements. For our scenario, the state vector of $\mathbf{A_1}$ is represented as \( \mathbf{x}_k\)  and the associated uncertainty as \( \mathbf{P}_k\)  as described in Equation \ref{e1}.
\begin{align}
\begin{cases}
     \mathbf{x}_{k} &= \begin{bmatrix}  x_1^k \quad y_1^k \quad  v_{x_1}^k \quad v_{y_1}^k   \end{bmatrix}^\top \\
     \mathbf{x}_{k|k-1} &= \mathbf{F}_k\mathbf{x}_{k-1|k-1} + \mathbf{w}_k \\
    \mathbf{P}_{k|k-1} &= \mathbf{F}_k\mathbf{P}_{k-1|k-1}\mathbf{F}_k^\top + \mathbf{Q}_k  \\
     \mathbf{y}_{k} &= \mathbf{z}_k - \mathbf{H}_k\mathbf{x}_{k|k-1}  \\
    \mathbf{K}_{k} &= \mathbf{P}_{k|k-1} \mathbf{H}_k^\top(\mathbf{H}_k \mathbf{P}_{k|k-1} \mathbf{H}_k^\top + \mathbf{R}_k) ^{-1}  \\
    \mathbf{x}_{k|k} &= \mathbf{x}_{k|k-1} + \mathbf{K}_k\mathbf{y}_k \\
    \mathbf{P}_{k|k} &= (\mathbf{I} - \mathbf{K}_k\mathbf{H}_k) \mathbf{P}_{k|k-1}
\end{cases}
     \label{kf}
\end{align}
where $\mathbf{x}{k|k-1}$ is the predicted state at time $k$ given the state at time $k-1$, and $\mathbf{P}{k|k-1}$ is the covariance of the predicted state estimate. The terms $\mathbf{F}_k$, $\mathbf{w}_k$, $\mathbf{Q}_k$, $\mathbf{z}_k$, $\mathbf{y}_k$, $\mathbf{H}_k$, $\mathbf{R}_k$, and $\mathbf{K}_k$ represent the system model, process noise, process noise covariance matrix, actual measurement, measurement residual, observation matrix, measurement noise covariance matrix, and Kalman gain, respectively. Finally, $\mathbf{x}_{k|k}$ and $\mathbf{P}_{k|k}$ are the updated state and covariance.

To estimate the state of Robot $\mathbf{A_2}$, which lacks constant direct access to absolute positioning, the system relies on inter-vehicle range measurements and state estimates communicated from Robot $\mathbf{A_1}$. Although the UT is traditionally applied in nonlinear estimation settings, its utility in this context lies in its ability to deterministically sample sigma points that capture the statistical properties of multiple distributions. By propagating these sigma points through the nonlinear system model, the UT effectively combines the uncertainty in $\mathbf{A_1}$'s state, $\mathbf{A_2}$'s predicted state, and the inter-vehicle range measurement to produce a new estimate for $\mathbf{A_2}$'s state. This capability makes UT particularly suited for merging disparate sources of information in a principled manner. 
However, this process inherently involves a form of information reuse: the predicted state of $\mathbf{A_2}$, obtained from its motion model, is used to generate an updated estimate, directly fusing these estimates without accounting for unknown cross-correlations can lead to overconfident and inconsistent state estimates. To mitigate this, Covariance Intersection (CI) methods are used. In particular, Split Covariance Intersection (SCI) assumes an unknown correlation between parts of the fused information and enforces conservative consistency by ensuring that the resulting covariance upper bounds the fused covariance.



\subsection{Unscented Transform}
The Unscented Transform (UT) \cite{unscentedtransform2004} is a mathematical method utilized to estimate the statistics (mean and covariance) of a random variable that undergoes a nonlinear transformation. The UT works by selecting a set of sigma points that capture the mean and covariance of the distribution, propagating them through the nonlinear function, and then calculating the statistics of the nonlinearly transformed mean and covariance by taking a weighted average of the transformed sigma points. 
\begin{align}
    &\mathcal{X}_{0} = \mathbf{x}, \quad  
    \mathcal{X}_{i} = \mathbf{x} + \left( \sqrt{(n + \lambda) \mathbf{P}} \right)_i, \notag \\
    &\mathcal{X}_{i+n} = \mathbf{x} - \left( \sqrt{(n + \lambda) \mathbf{P}} \right)_i
    \label{sig} \\
    &\mathcal{Y}_i = f_{nl}(\mathcal{X}_i), \quad i = 0, \dots, 2n
    \label{nl} \\
    &W_{0}^{m} = \frac{\lambda}{n +\lambda }, \quad  
    W_{0}^{c} = \frac{\lambda}{n + \lambda } + (1-\alpha ^{2}+\beta), \notag \\
    &W_{i}^{m} = W_{i}^{c} = \frac {1}{2(n+\lambda)}, \quad i = \{1,2,\ldots,2n\}
    \label{wgh} \\
    &\mathbf{y} =\sum _{i=0}^{2n}W_{i}^{m}{\mathcal {Y}_i}, \notag \\
    &\mathbf{P}_y  =\sum _{i=0}^{2n}W_{i}^{c}(\mathcal{Y}_i - \mathbf{y}) (\mathcal{Y}_i - \mathbf{y})^\top
    \label{utmc}
\end{align}

For the sigma points selection step, $\mathcal{X}_{0}$ is the mean state vector, while $\mathcal{X}_{i}$ and $\mathcal{X}_{i+n}$ are the sigma points, distributed symmetrically around the mean. The mean and covariance are denoted by $\mathbf{x}$ and $\mathbf{P}$, respectively. $n$ represents the dimension of the state, and $\lambda$ is a scaling parameter. From Equation \ref{sig}, each sigma point $\mathcal{X}_{i}$ will be propagated through the nonlinear function as shown in Equation \ref{nl}.                         
Accordingly, to provide an accurate representation of the mean and covariance, the set of transformed sigma points is weighted, and the weights can be derived according to Equation \ref{wgh} where $\alpha$ and $\beta$ are the scaling parameters. The estimated mean and covariance are then obtained by computing Equations \ref{utmc} 

In this work, the UT specifically addresses the problem of determining the resultant distribution of $\mathbf{A_2} $'s state after it is connected through a nonlinear function $f_{nl}(.)$  to the state of $\mathbf{A_1}$ whose distribution is known. UT is essential to the functionality of the Split Covariance Intersection (SCI) method, which requires an estimate of $\mathbf{A_2}$ directly, as it accurately handles nonlinearities in state estimation and covariance propagation and provides a mechanism to do so. The UT is explicitly utilized to compute one of the state estimates and its corresponding covariance, which are used in the SCI method to ensure that all sources of uncertainty are conservatively accounted for in the fusion process.
\subsection{Split Covariance Intersection}
Split Covariance Intersection (SCI) \cite{li2013split} is a data fusion algorithm that handles correlated measurements under unknown correlations. SCI is an extension of the Covariance Intersection method, which is widely used for conservative fusion in distributed systems. Unlike standard Covariance Intersection (CI), SCI splits the estimation process into two parts: a mutually correlated component and an independent component. By splitting the covariance matrices, the SCI can fuse the estimates more precisely while avoiding double-counting of correlated uncertainties.
Given two estimates ($\mathbf{x}_1$, $\mathbf{P}_1$) and ($\mathbf{x}_2$, $\mathbf{P}_2$), SCI decomposes their covariances into:
\begin{align}
\mathbf{P}_1 &= \mathbf{P}_1^d + \mathbf{P}_1^i \\
\mathbf{P}_2 &= \mathbf{P}_2^d + \mathbf{P}_2^i
\end{align}
where superscripts $d$ and $i$ denote the correlated and independent components, respectively. The fused estimate is then computed as:
\begin{align}
\begin{cases}
    \mathbf{P}_1 &= \mathbf{P}_1^d/w + \mathbf{P}_1^i \\
\mathbf{P}_2 &= \mathbf{P}_2^d/(1-w) + \mathbf{P}_2^i \\
\mathbf{P}^{-1} &= \mathbf{P}_1^{-1} + \mathbf{P}_2^{-1}\\
\mathbf{X} &= \mathbf{P} ( \mathbf{P}_1^{-1}\mathbf{X}_1 + \mathbf{P}_2^{-1}\mathbf{X}_2 ) \\
\mathbf{P}^i &= \mathbf{P} ( \mathbf{P}_1^{-1}\mathbf{P}_1^i \mathbf{P}_1^{-1} + \mathbf{P}_2^{-1}\mathbf{P}_2^i \mathbf{P}_2^{-1} )\mathbf{P} \\
\mathbf{P}^d &= \mathbf{P}- \mathbf{P}_i
\end{cases}
\end{align}
where $\omega \in [0,1]$ is a weighting parameter that is typically chosen to minimize the trace or determinant of $\mathbf{P}$.
The independent components are fused using the standard Kalman filter update, while the correlated components are fused using CI. This split approach results in tighter covariance bounds compared to standard CI, more accurate state estimates, and guaranteed consistency when the decomposition is conservative. 

SCI is particularly useful in this work because it enables robust and consistent cooperative localization when agents share uncertain state estimates without knowing their exact cross-correlations, unlike traditional methods such as Kalman Filters, which require explicit knowledge of cross-correlations. This is particularly useful in this application due to the known information reuse employed in the UT.

In this work, SCI is used in conjunction with UT to provide a proposed formulation designed to handle nonlinear state estimation and conservative fusion. The UT is applied to propagate the uncertainty of one agent's state through a nonlinear transformation, providing an estimated distribution for another agent based on shared range measurements, which introduces a set of unmodeled cross-correlations between the estimates. Since the resulting estimate maintains dependence on the agent's local state estimate, SCI is then used to fuse the two in a way that avoids overconfidence while benefiting from shared information. 


The key challenge in SCI lies in determining the split between correlated and independent components. A common approach is to use domain knowledge or conservative bounds.

\subsection{Unscented Transform with Split Covariance Intersection}
To achieve the objectives defined in the introductory section, an integrated approach that combines the Unscented Transform (UT) and Split Covariance Intersection (SCI) techniques is proposed to address the challenges of nonlinearity and uncertainty in state estimation. The primary objective is to develop a robust localization framework that can effectively fuse measurements and state information from two vehicles operating under different sensor constraints. 

Given a scenario with two agents named Robot 1 and Robot 2, which will be referred to as $\mathbf{A_1}$ and $\mathbf{A_2}$ in this formulation. $\mathbf{A_1}$ is assumed to have the availability of frequent positioning updates with minimal noise. Thus, its estimate of its position is more reliable with minimized uncertainty. $\mathbf{A_2}$, on the other hand, gets sporadic and noisy positioning updates, which makes its estimation unreliable with a larger uncertainty. The goal of this method is to fuse information shared between these agents to improve $\mathbf{A_2} $'s state estimate, using only range measurements. The main challenge lies in the fact that only range measurements are available between the two agents without bearing or directional information. This method is particularly valuable in scenarios where the spatial relationship between agents is uncertain and traditional covariance intersection methods, which typically rely on range and bearing measurements, cannot be directly applied. 

Recall that $\mathbf{A_1}$ gets reliable measurements; thus, its state can be estimated using a simple Kalman Filter formulation. The Unscented Transform and Split Covariance Intersection fusion technique is used to estimate the state of $\mathbf{A_2}$; the architecture of the method is described in Figure \ref{archi} below.
\begin{figure}[htb!]
    \centering
    \begin{tikzpicture}
        
            \draw[thick] (2, 9) ellipse (1.5 and 0.5);
            \node at (2, 9) {Start/Initialization};
        
            \draw[thick] (-.1, 7) rectangle (4.1, 8);
            \node at (2, 7.5) {State Prediction};

            \node[diamond, draw, thick, inner sep=1pt, aspect=4] at (2, 6) {Check Pos Update};
            \node at (5, 6.3) {YES};
            \node at (2.6, 5.3) {NO};

            \draw[thick] (4.3, 4) rectangle (7.5, 5);
            \node[text width=4.7cm, align=center] at (5.9, 4.5) {Measurement Update};
            \draw[->, thick] (4.1, 6) -- (5.8, 6) -- (5.8, 5) ; 

            \draw[thick] (4.3, 2.5) rectangle (7.5, 3.5);
            \node[text width=4.7cm, align=center] at (5.9, 3) {Estimate State};
            \draw[->, thick] (5.8, 4) -- (5.8, 3.5) ; 
             \draw[->, thick] (5.8, 2.5) -- (5.8, 1.5) -- (4.1 , 1.5) ; 

            
        
            \draw[thick] (-0.1, 4) rectangle (4.1, 5);
            \node[text width=4.7cm, align=center] at (2, 4.5) {Apply Unscented Transform for New State Prediction};
        
            \draw[thick] (-0.1, 2.5) rectangle (4.1, 3.5);
            \node[text width= 5.5cm, align=center] at (2, 3)  {Split Covariance Intersection};

            \draw[thick] (-0.1, 1) rectangle (4.1, 2);
            \node[text width= 5.5cm, align=center] at (2, 1.5)  {Publish State and Covariance};
        
            \draw[->, thick] (2, 8.5) -- (2, 8); 
            \draw[->, thick] (2, 7) -- (2, 6.5);     
            \draw[->, thick] (2, 5.5) -- (2, 5); 
            \draw[->, thick] (2, 4) -- (2, 3.5);     
            \draw[->, thick] (2, 2.5) -- (2, 2);     

            \draw[->, thick] (-.1, 1.5) -- (-.5, 1.5) -- (-.5, 7.5) -- (-.1, 7.5); 
        \end{tikzpicture}
    \caption{Architecture of the Proposed method }
    \label{archi}
\end{figure}
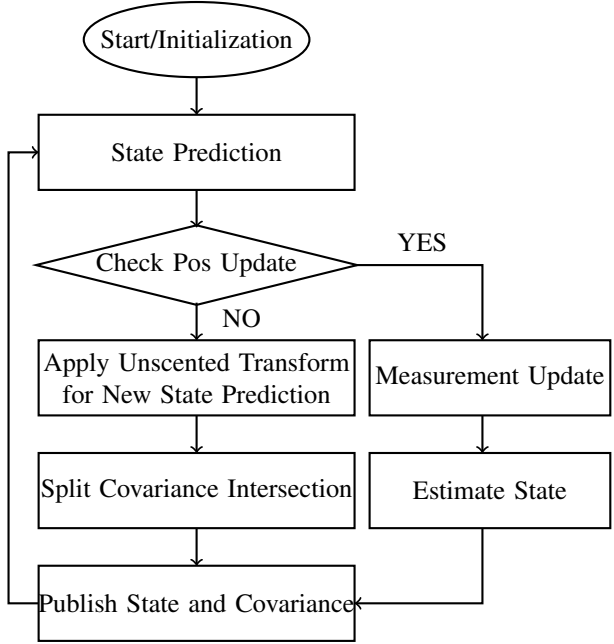
The state of $A_2$ is originally initialized to set values that are defined based on prior knowledge and assumptions about the system. In the prediction stage, the state of the system is projected forward in time using the system's dynamic model. The state transition equation is given by:
\begin{align}
    \mathbf{\hat{x}}_k^{A_2} = f(\mathbf{x}_{k-1}, \mathbf{u}_{k-1}) + \mathbf{w}_{k},
\end{align}
where $\mathbf{\hat{x}}_k^{A_2}$ is the predicted state at time $k$ based on the utilized motion model, $f$ is the state transition function, $\mathbf{u}_{k-1}$ is the control input, and $\mathbf{w}_k$ is the process noise assumed to be zero-mean Gaussian noise. The next step is to check for the availability of a new positioning measurement. If a positioning update is available, the measurement update step and the state can be further estimated using the Kalman Filter formulation. In the case where there is no positioning update, the system proceeds to apply the UT for another prediction based on the state and range measurements received from $A_1$.

The UT is used specifically in this formulation to make a prediction for the state of $A_2$ based on range information. This prediction can then be used by the SCI for fusion. The UT will handle the nonlinearity that arises from the range measurement. Because the UT is utilized to make an estimate for the position of $A_2$ by incorporating the range measurement from the UWB, the input to the UT module will be an augmented state and covariance matrix shown below

\begin{align*}
x_{AUG} = 
    \begin{bmatrix}
        \mathbf{x}^{A_1} \\ \mathbf{\hat{x}}_k^{A_2} \\ \mathbf{r}
    \end{bmatrix} , \quad  
    P_{AUG} = 
    \begin{bmatrix}
        P^{A_1}_{2 \times 2}  & 0_{2 \times 2} & 0_{2 \times 1} \\
        0_{2 \times 2}  & P^{A_2}_{2 \times 2} & 0_{2 \times 1} \\
        0_{1 \times 2}  & 0_{1 \times 2} & P^{r}_{1 \times 1} 
    \end{bmatrix}
\end{align*}

The sigma points $\mathcal{X}$ will be created using Equation \ref{sig}. The transformation applied to these sigma points utilizes the geometric relationship between the two robots: it assumes that the position of $\mathbf{A_2}$ lies along the direction vector defined by the relative pose between $\mathbf{A_2}$ and $\mathbf{A_1}$, scaled by the sampled range $r_k$. Formally, the transformation function $\mathcal{T}(\cdot)$ applied to each sigma point computes the direction vector $\mathbf{u}$ and a relative distance $\mathbf{d}$.
\begin{align}
    \mathbf{u} &= \mathcal{X}_i^{A_2} - \mathcal{X}_i^{A_1} \\
    \mathbf{d} &= \| \mathcal{X}_i^{A_2} - \mathcal{X}_i^{A_1} \| \\
     \mathcal{Y} &= \mathcal{T}(\mathcal{X}_i^{A_1},\mathcal{X}_i^{A_2},\mathcal{X}_i^{r}) = \mathcal{X}_i^{A_2}  + \mathcal{X}_i^{r} + \frac{\mathbf{u}}{\mathbf{d}}
\end{align}
a new positional estimate of $\mathbf{A_2}$, $\mathbf{x}^{*A_2}$ and associated uncertainty $\mathbf{P}^{*A_2}$    is computed using Equations \ref{wm} and \ref{wc}.;
\begin{align}
\mathbf{x}^{*A_2} &= \sum_{i=0}^{2n} w_i \mathcal{Y}_{(i)} \label{wm}\\
 \mathbf{P}^{*A_2} &= \sum_{i=0}^{2n} w_i (\mathcal{Y}_{(i)} - \mathbf{x}^{*A_2})(\mathcal{Y}_{(i)} - \mathbf{x}^{*A_2})^\top \label{wc}
\end{align}
In order to use the SCI method to fuse the states $\mathbf{\hat{x}}^{A_2}$ and $\mathbf{x}^{*A_2}$,  the corresponding covariances  $\mathbf{\hat{P}}^{A_2}$ and $\mathbf{P}^{*A_2}$ are decomposed into their dependent and independent components. 

+Let \(\mathbf{P}_{\text{E}} = \mathbf{\hat{P}}^{A_2},  \quad \mathbf{P}_{\text{E}}^* = \mathbf{P}^{*A_2}, \quad \mathbf{x}_{\text{E}} = \mathbf{\hat{x}}^{A_2}, \quad \mathbf{x}_{\text{E}}^* = \mathbf{x}^{*A_2}\).
\begin{gather}
    \mathbf{P}_{\text{E}} =  \mathbf{P}_{dE(t)} + \mathbf{P}_{iE(t)}\\
\mathbf{P}_{\text{E}}^* =  \mathbf{P}_{dE(t)}^* + \mathbf{P}_{iE(t)}^*
\end{gather} where subscript $d$ denotes the dependency of the covariance matrix of the estimated state with respect to the other state, and subscript $i$ denotes the independence of the covariance matrix of the estimated state with respect to the other state. \\
For example, initially, $\mathbf{P}_{dE(t=0)} =0$ because it has no dependency since the other state and covariance ($P_E^*$) does not influence the initial calculation of the $P_E$, also, $\mathbf{P}_{iE(t)}$ is initially fully independent from $P_E^*$. Conversely, $\mathbf{P}_{dE(t=0)}^*$ is initially fully dependent on $P_E$ since it is calculated with UT using $P_E$, and it is not independent from $P_E$ in any time (e.g., not initially nor after the relative update) which means $\mathbf{P}_{iE(t)}^* =0$.   
Using the split covariance intersection as follows,
\begin{align}
\begin{cases}
   \mathbf{P}_1 &= \mathbf{P}_{dE(t)}/w + \mathbf{P}_{iE(t)} \\
\mathbf{P}_2 &= \mathbf{P}_{dE(t)}^*/(1-w) + \mathbf{P}_{iE(t)}^* \\
\mathbf{K} &= \mathbf{P}_1(\mathbf{P}_1 + \mathbf{P}_2)^{-1} \\
\mathbf{X}_{E(t)} &= \mathbf{X}_{E(t)} + \mathbf{K}(\mathbf{X}_{E(t)}^* - \mathbf{X}_{E(t)}) \\
\mathbf{P}_{E(t)} &= (\mathbf{I} - \mathbf{K})\mathbf{P}_1 \\
\mathbf{P}_{iE(t)} &= (\mathbf{I} - \mathbf{K})\mathbf{P}_{iE(t)}(\mathbf{I} - \mathbf{K})^T + \mathbf{K}\mathbf{P}_{iE(t)}^*\mathbf{K}^T \\
\mathbf{P}_{dE(t)} &= \mathbf{P}_{E(t)} - \mathbf{P}_{iE(t)}. 
\end{cases}
    \label{sci}
\end{align}
Again, the assumption for the initial case to use the equation set from \ref{sci} and the assumption for the independent decomposed value of the $\mathbf{P}_{E(t)}^*$ are
\begin{align}
    \mathbf{P}_{dE(t=0)} = 0 \quad
    \mathbf{P}_{iE(t)}^* = 0  
\end{align}
Therefore, initially, the equations \ref{sci}a and b are assumed to become
\begin{align}
    \mathbf{P}_1 = \mathbf{P}_{iE(t=0)}, \quad
\mathbf{P}_2 = \mathbf{P}_{dE(t=0)}^*/(1-w) 
\end{align}
where $w \in [0,1]$ is initially assumed as $w\approx0$. The calculation of the $w$ is done by minimizing the determinant of the new covariance. The rest of Equation \ref{sci} is computed and the $\mathbf{X}_{E(t)}, \mathbf{P}_{E(t)}, \mathbf{P}_{iE(t)}$ and $\mathbf{P}_{dE(t)}$  are published.
For each measurement update (i.e., positioning update), $\mathbf{P}$ and the $\mathbf{P}_{iE}$ need to be updated accordingly, and the Joseph form is used for both cases to ensure numerical stability and maintain the positive semi-definiteness of the covariance matrices. For example, when we have a positioning update, we have this covariance update equation:
\begin{align}
    \mathbf{P_E} &= (\mathbf{I}_{15} - \mathbf{K}_{pos} \mathbf{H}_{pos}) \mathbf{P_{iE}} (\mathbf{I}_{15} - \mathbf{K}_{pos} \mathbf{H}_{pos})^\top + \ldots  
    \notag \\ &\qquad \mathbf{K}_{pos} \mathbf{R}_{pos} \mathbf{K}_{pos}^\top 
\end{align}
Similarly, $P_{iE}$ is updated using the same formula.
\begin{align}
    \mathbf{P_{iE}} &= (\mathbf{I}_{15} - \mathbf{K}_{pos} \mathbf{H}_{pos}) \mathbf{P_{iE}} (\mathbf{I}_{15} - \mathbf{K}_{pos} \mathbf{H}_{pos})^\top + \ldots  \notag \\
    &\qquad  \mathbf{K}_{pos} \mathbf{R}_{pos} \mathbf{K}_{pos}^\top \\
     \mathbf{P_{dE}}& = \mathbf{P_{E}} - \mathbf{P_{iE}}
\end{align}
Then, the updated $\mathbf{P_{iE}}$, $\mathbf{P_{E}}$, and $\mathbf{P_{dE}}$ as well  $\mathbf{X_{E}}$ will be published, and the process will continue with Equations \ref{sci}, but this time, only $\mathbf{P}_{iE(t)}^*$ will be 0.

\section{Experimental Design}
\label{experimental_results}
In this experimental design,  the efficacy of the proposed cooperative localization method is evaluated by extensive experimental evaluations using a two-dimensional simulation framework with a simplified scenario. The simulation utilizes a constant velocity motion model for robot dynamics, where range measurements are calculated based on the Euclidean distance between two robots. Noise in this experiment refers to an additive Gaussian error term incorporated into the range measurements and positioning measurement updates to reflect real-world sensor imperfections. Ground-truth position updates are provided based on the actual positions of the robots, serving as a reference for performance assessment.

\begin{figure*}[h]
    \centering
    \begin{minipage}{0.32\textwidth}
        \centering
        \includegraphics[width=0.95\linewidth]{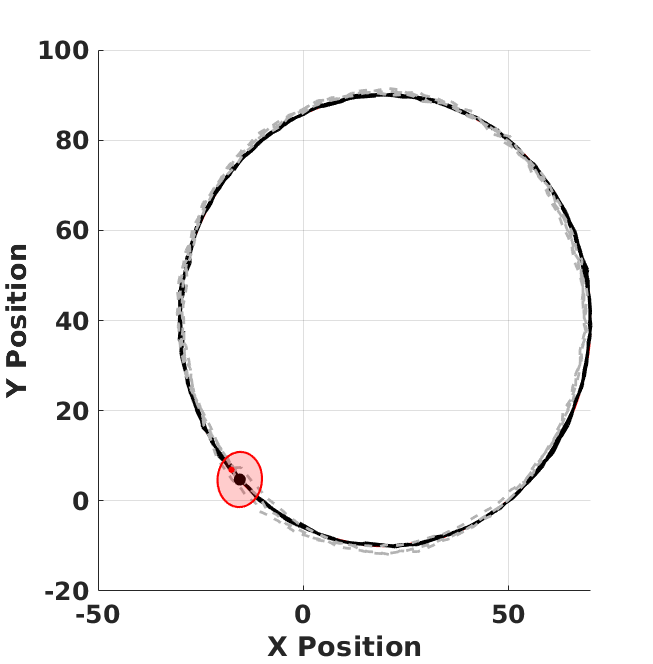}
    \end{minipage}%
    \begin{minipage}{0.32\textwidth}
        \centering
        \includegraphics[width=0.95\linewidth]{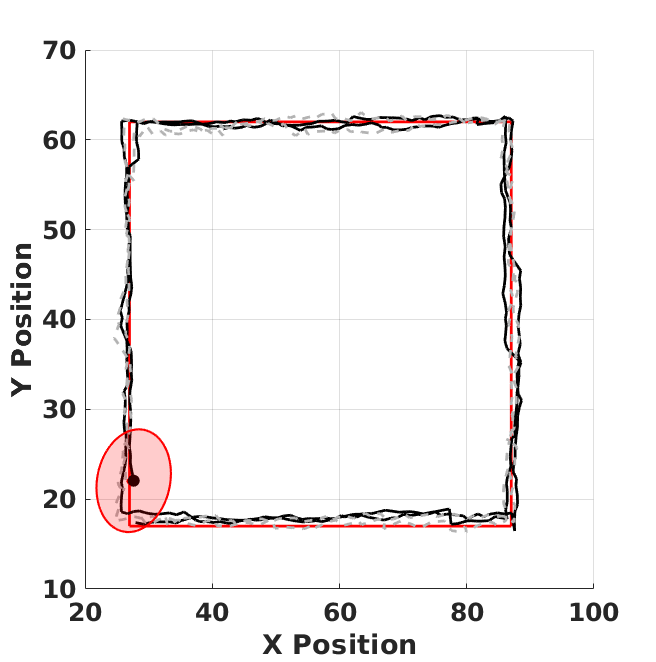}
    \end{minipage}%
    \begin{minipage}{0.32\textwidth}
        \centering
        \includegraphics[width=0.95\linewidth]{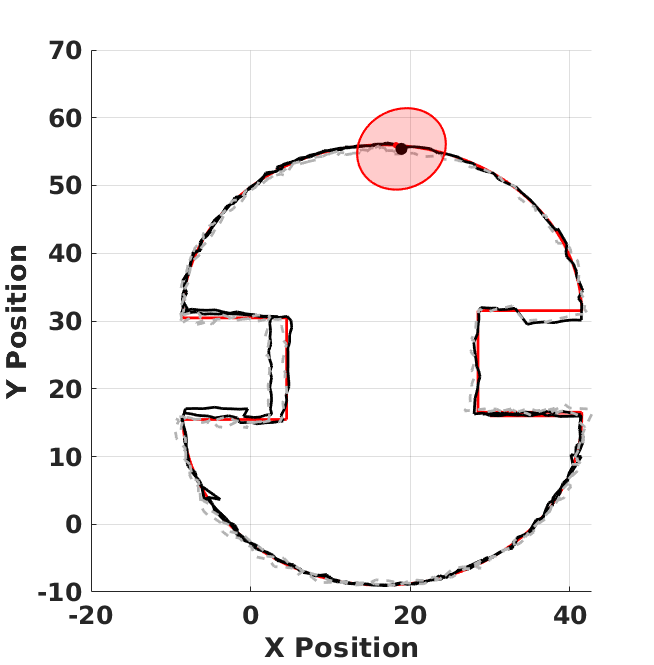}
    \end{minipage}
    \caption{The three distinct paths taken by Robot 2. The red line represents the true position of the robot, and the black line indicates the estimate obtained using the proposed method. The red ellipse is the uncertainty around the estimated position.}
    \label{paths}
\end{figure*}

Using Monte Carlo simulations with varying noise values and initial conditions,  a sensitivity analysis is performed to further evaluate the proposed method's robustness and reliability. This enables the assessment of the method's reliability across multiple iterations and evaluates its sensitivity to various operating conditions and measurement uncertainties. This approach provides insights into the method's effectiveness and potential limitations in practical scenarios.\\
The simulation environment is set within a 200 by 200 unit area, and the starting positions of the robots are randomly initialized for each run to account for varying initial conditions. In the 2D simulation, the first robot moves along a rectangular path, while the second robot, whose state is being estimated by the proposed method, can travel along three distinct paths, as shown in Figures \ref{paths} $a$, $b$, and $c$.

The table below summarizes the key parameters used in the 2D simulation and Monte Carlo analysis. These parameters include the motion model characteristics, noise levels for range and positioning measurements, and Monte Carlo Simulation settings.

\noindent
\begin{table}[htb!]
    \centering
    \caption{Parameters for the Monte Carlo Simulation}
    \renewcommand{\arraystretch}{1.2}
    \setlength{\tabcolsep}{6pt}
    \begin{tabular}{l c c}
        \toprule
        \textbf{Parameter} & \textbf{Value} & \\ 
        \midrule
        Range Noise & Mean: 0, Std: 5 & \\ 
        Positioning Noise & Mean: 0, Std: 3 & \\  
        Velocity Offset & Mean: 0, Std: 1 & \\ 
        Position Offset & 0 - 10 & \\ 
        Path & Circle, Rectangle, Donut & \\ 
        Number of Runs & 40000 & \\ 
        Epochs per Run & 400 & \\ 
        \bottomrule
    \end{tabular}
    
    \label{param}
\end{table}

\section{Results}
\label{results}
The results of the analysis of the Monte Carlo simulation are presented in this section. We investigated the localization accuracy of the approach by computing the Euclidean distance between the estimated position and the true position, and obtained the corresponding statistics from the data. Table \ref{tab:localization_metrics} summarizes the RMS error results for runs compared by the paths. 

\begin{table}[h!]
    \centering
    \caption{Localization Performance Metrics for Range Noise}
    \renewcommand{\arraystretch}{1.2}  
    \setlength{\tabcolsep}{6pt}  
    \resizebox{\columnwidth}{!}{  
    \begin{tabular}{l c c c c c c c}
        \toprule
        \textbf{Metric} & $\mathbf{0\sigma}$ & $\mathbf{1/2\sigma}$ & $\mathbf{\sigma}$ & $\mathbf{3/2\sigma}$ & $\mathbf{2\sigma}$ & $\mathbf{5/2\sigma}$ & $\mathbf{3\sigma}$ \\
        \midrule
        NumRuns & 15347 & 11940 & 7394 & 3519 & 1308 & 390 & 102 \\
        Mean Error (m) & 3.65 & 3.78 & 3.96 & 4.25 & 4.78 & 5.24 & 5.91 \\
        Std Error (m) & 1.80 & 1.78 & 1.79 & 1.89 & 2.23 & 2.33 & 2.58 \\
        Median Error (m) & 3.16 & 3.29 & 3.48 & 3.72 & 4.09 & 4.52 & 5.13 \\
        Max Error (m) & 16.15 & 17.15 & 17.31 & 19.72 & 24.65 & 16.01 & 14.82 \\
        Min Error (m) & 0.83 & 1.07 & 1.08 & 1.18 & 1.37 & 1.62 & 1.65 \\
        \bottomrule
    \end{tabular}
    }
    \label{tab:uwbvsrmse}
\end{table}

\begin{figure}[h]
    \centering
    \includegraphics[width=1\linewidth]{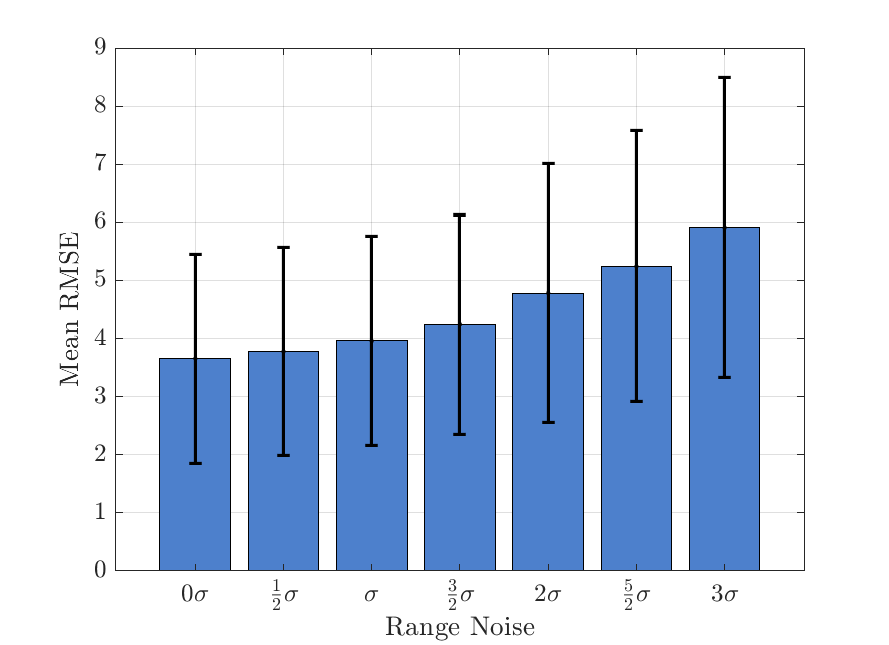}
    \caption{Effect of Range Variance on RMSE}
    \label{uwbvsrmse}
\end{figure}

Figure \ref{uwbvsrmse} and Table \ref{tab:uwbvsrmse} illustrate the impact of increasing range noise on localization accuracy, measured by the Root Mean Square Error (RMSE). As the range noise level increases from $0\sigma$ to $3\sigma$, the mean RMSE exhibits a clear upward trend, rising from 3.65 meters to 5.91 meters. This indicates a direct degradation in localization performance with higher measurement uncertainty. Additionally, the standard deviation of the error, as seen from the error bars, also increases, reflecting greater variability in the estimates. Notably, the rise in both mean and variability becomes pronounced beyond $1.5\sigma$, suggesting a nonlinear sensitivity to noise. 


\begin{table}[h!]
    \centering
    \caption{Localization Performance Metrics for Pos Noise}
    \renewcommand{\arraystretch}{1.2}  
    \setlength{\tabcolsep}{6pt}  
    \resizebox{\columnwidth}{!}{  
    \begin{tabular}{l c c c c c c c}
        \toprule
        \textbf{Metric} & $\mathbf{0\sigma}$ & $\mathbf{1/2\sigma}$ & $\mathbf{\sigma}$ & $\mathbf{3/2\sigma}$ & $\mathbf{2\sigma}$ & $\mathbf{5/2\sigma}$ & $\mathbf{3\sigma}$ \\
        \midrule
        NumRuns & 15364 & 11944 & 7408 & 3488 & 1279 & 418 & 99 \\
        Mean Error (m) & 2.57 & 3.51 & 4.83 & 6.29 & 7.95 & 9.39 & 10.94 \\
        Std Error (m) & 0.88 & 0.88 & 1.09 & 1.41 & 1.79 & 2.00 & 2.20 \\
        Median Error (m) & 2.69 & 3.38 & 4.63 & 6.02 & 7.65 & 9.16 & 10.75 \\
        Max Error (m) & 19.72 & 24.65 & 16.01 & 17.03 & 16.94 & 17.31 & 17.15 \\
        Min Error (m) & 0.83 & 1.56 & 2.29 & 3.36 & 4.13 & 5.45 & 6.62 \\
        \bottomrule
    \end{tabular}
    }
    \label{tab:gpsvsrmse}
\end{table}

\begin{figure}[h!]
    \centering
    \includegraphics[width=1\linewidth]{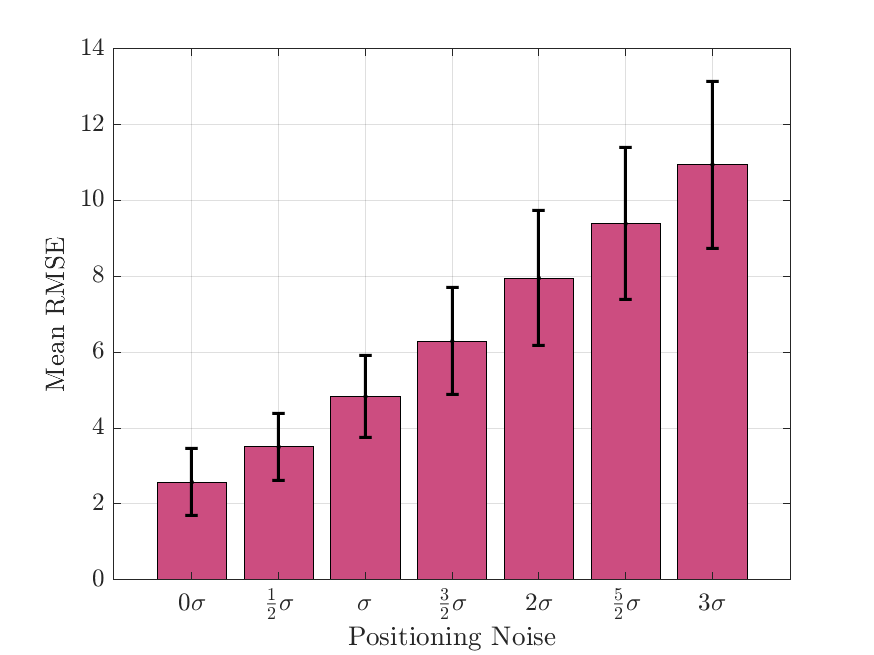}
    \caption{Effect of Positioning solution uncertainty on RMSE}
    \label{gpsvsrmse}
\end{figure}

The degradation in localization performance as positioning noise increases is illustrated in Figure \ref{gpsvsrmse} and summarized in Table \ref{tab:gpsvsrmse}. The bar chart shows a clear positive correlation between the level of injected noise, ranging from $0\sigma$ to $3\sigma$, and the mean root mean square error (RMSE). Starting at 2.57 meters for $0\sigma$, the RMSE rises steadily, reaching 10.94 meters at $3\sigma$. This trend is accompanied by a growing standard deviation, indicating increased uncertainty in the estimates. The table further quantifies this behavior with detailed statistics: the mean and median errors increase monotonically, while the minimum error shifts upward from 0.83 m to 6.62 m, indicating that even the best-case localization deteriorates with increased noise. These results clearly underscore the sensitivity of the localization algorithm to positioning noise and highlight the importance of accurate positioning inputs for maintaining reliable state estimates.
    
\begin{table}[h!]
    \centering
    \caption{Localization Performance Metrics for Initial Position Offset}
    \renewcommand{\arraystretch}{1.2}  
    \setlength{\tabcolsep}{6pt}  
    \resizebox{\columnwidth}{!}{  
        \begin{tabular}{l c c c c c c c c c c c}
        \toprule
        \textbf{Metric} & \textbf{0} & \textbf{1} & \textbf{2} & \textbf{3} & \textbf{4} & \textbf{5} & \textbf{6} & \textbf{7} & \textbf{8} & \textbf{9} & \textbf{10} \\
        \midrule
        NumRuns & 3585 & 3627 & 3705 & 3655 & 3756 & 3592 & 3557 & 3579 & 3688 & 3649 & 3607 \\
        Mean Error (m) & 3.62 & 3.68 & 3.63 & 3.71 & 3.83 & 3.79 & 3.90 & 3.93 & 4.02 & 4.11 & 4.22 \\
        Std Error (m) & 1.87 & 1.97 & 1.86 & 1.85 & 1.91 & 1.85 & 1.83 & 1.79 & 1.77 & 1.76 & 1.77 \\
        Median Error (m) & 3.16 & 3.20 & 3.17 & 3.27 & 3.28 & 3.26 & 3.33 & 3.38 & 3.47 & 3.57 & 3.66 \\
        Max Error (m) & 19.72 & 15.19 & 16.94 & 24.65 & 14.82 & 14.77 & 14.79 & 15.06 & 17.31 & 17.88 & 14.34 \\
        Min Error (m) & 0.83 & 0.90 & 1.07 & 1.13 & 1.29 & 1.39 & 1.59 & 1.67 & 1.92 & 2.04 & 2.23 \\
        \bottomrule
    \end{tabular}
    }
    \label{tab:povsrmse}
\end{table}

\begin{figure}[h!]
    \centering
    \includegraphics[width=1\linewidth]{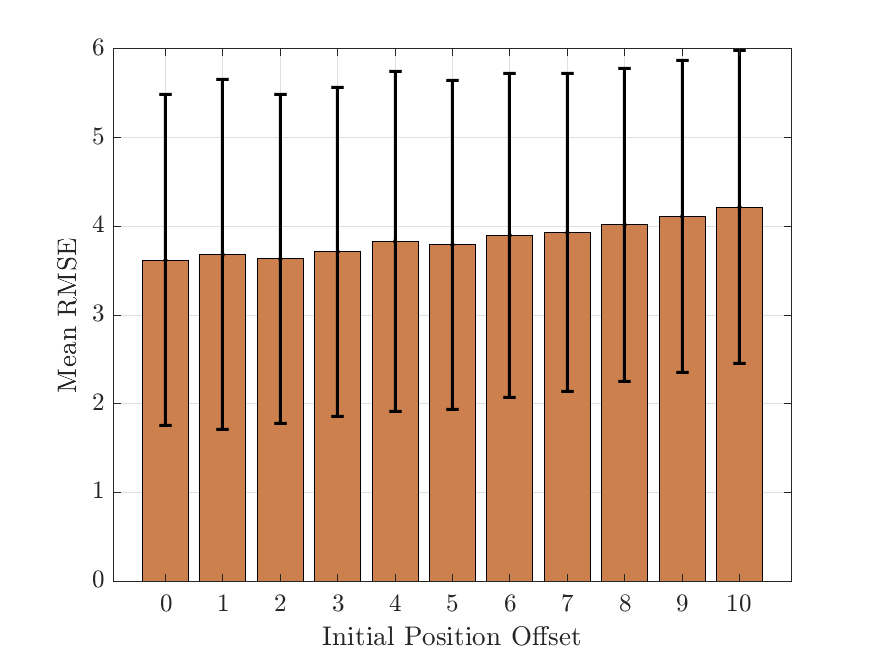}
    \caption{Effect of Initial Position Offset on RMSE}
    \label{povsrmse}
\end{figure}

The sensitivity of the localization algorithm to initial position offset is examined in Figure \ref{povsrmse} and Table \ref{tab:povsrmse}. As the offset increases from 0 to 10 meters, a gradual upward trend in the mean RMSE is observed, rising from 3.62 meters to 4.22 meters. This increase in error suggests that while the algorithm maintains a level of robustness to small initial inaccuracies, larger offsets introduce greater localization error. Interestingly, the standard deviation remains relatively stable, fluctuating only slightly between 1.76 and 1.97 meters, indicating consistent variability in estimator performance across the range of offsets. Additionally, the minimum error increases from 0.83 meters to 2.23 meters, indicating a decline in precision in favorable scenarios as offsets increase. While accurate initialization improves localization accuracy, the results show that the estimator remains relatively robust even under substantial initialization errors, indicating that accurate initialization is beneficial but not critical to overall performance.


\begin{figure}[h!]
    \centering
    \includegraphics[width=1\linewidth]{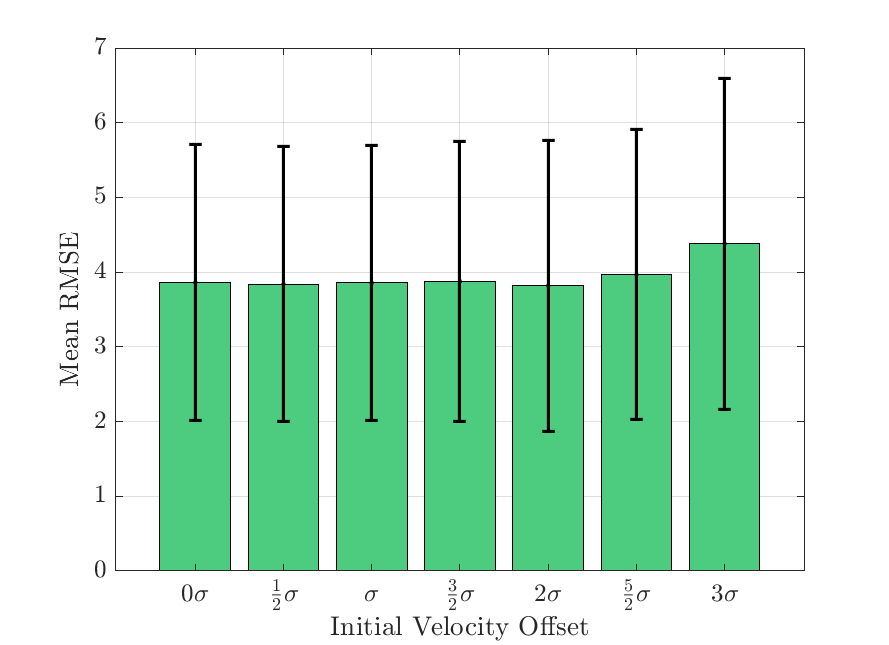}
    \caption{Effect of Initial Velocity Offset on RMSE}
    \label{vovsrmse}
\end{figure}

Table~\ref{tab:vovsrmse} presents the localization performance metrics for various levels of initial velocity offset, while the overall trend is further illustrated in Figure~\ref{vovsrmse}. Despite increasing initialization errors, the filter exhibits strong robustness, with the mean and median velocity errors remaining largely stable up to $2\sigma$ offsets and only moderate growth in standard deviation. Even at higher perturbation levels ($2.5\sigma$ and $3\sigma$), the increase in error remains bounded. These trends imply that while accurate velocity initialization can slightly improve consistency, it is not a critical requirement for achieving reliable localization performance. The estimator demonstrates resilience to moderate initialization errors, reinforcing the robustness of the proposed method.


\begin{table}[h!]
    \centering
    \caption{Localization Performance Metrics for Initial Velocity Offset}
    \renewcommand{\arraystretch}{1.2}  
    \setlength{\tabcolsep}{6pt}  
    \resizebox{\columnwidth}{!}{  
    \begin{tabular}{l c c c c c c c}
        \toprule
        \textbf{Metric} & $\mathbf{0\sigma}$ & $\mathbf{1/2\sigma}$ & $\mathbf{\sigma}$ & $\mathbf{3/2\sigma}$ & $\mathbf{2\sigma}$ & $\mathbf{5/2\sigma}$ & $\mathbf{3\sigma}$ \\
        \midrule
        NumRuns & 15256 & 11972 & 7406 & 3555 & 1284 & 395 & 132 \\
        Mean Error (m/s) & 3.86 & 3.84 & 3.86 & 3.88 & 3.82 & 3.96 & 4.38 \\
        Std Error (m/s) & 1.84 & 1.84 & 1.84 & 1.87 & 1.94 & 1.94 & 2.22 \\
        Median Error (m/s) & 3.35 & 3.37 & 3.34 & 3.34 & 3.27 & 3.34 & 3.86 \\
        Max Error (m/s) & 24.65 & 17.03 & 16.94 & 19.72 & 17.88 & 12.64 & 11.59 \\
        Min Error (m/s) & 0.90 & 0.83 & 1.00 & 0.90 & 1.12 & 1.22 & 1.06 \\
        \bottomrule
    \end{tabular}
    }
    \label{tab:vovsrmse}
\end{table}

\begin{figure}
    \centering
    \includegraphics[width=1\linewidth]{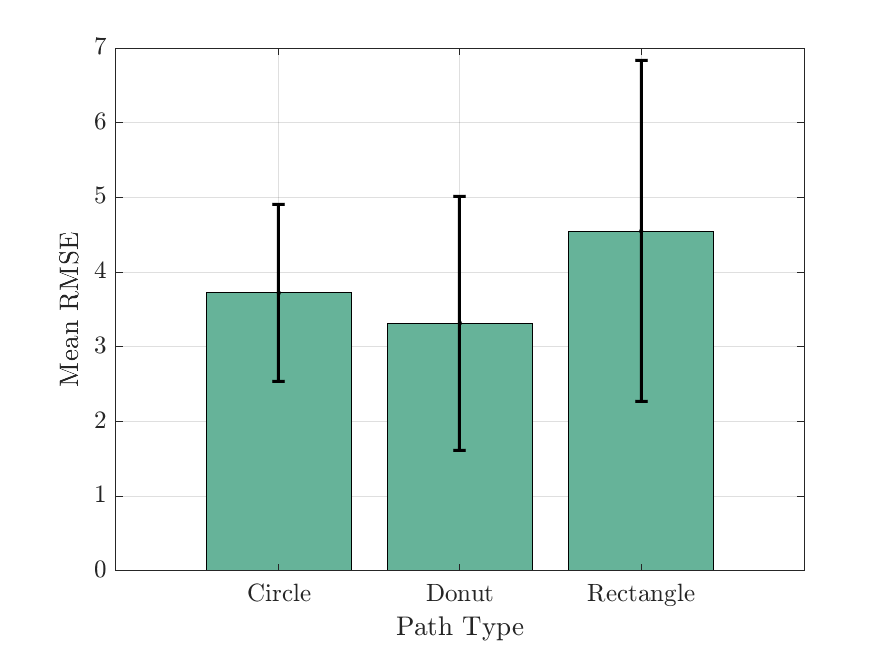}
    \caption{Effect of Path on RMSE}
    \label{pathvsrmse}
\end{figure}

Figure~\ref{pathvsrmse} and Table~\ref{tab:localization_metrics} evaluate the effect of trajectory geometry on localization accuracy. Results show that the donut path yields the lowest mean and median errors, indicating that moderate maneuvering with continuous heading variation supports better observability. In contrast, the rectangle path, despite its structured layout, suffers from the highest mean and standard deviation of errors, likely due to extended linear segments limiting observability. The circle path shows intermediate performance, benefiting from consistent motion but offering less excitation compared to the donut. 

\vspace{-10pt}
\noindent
\begin{table}[h!]
    \centering
    \caption{Localization Performance Metrics for Paths}
    \renewcommand{\arraystretch}{1.2}
    \setlength{\tabcolsep}{6pt}
    \begin{tabular}{l c c c}
        \toprule
        \textbf{Metric} & \textbf{Circle} & \textbf{Donut} & \textbf{Rectangle} \\ 
        \midrule
        NumRuns & 13327 & 13491 & 13182 \\ 
        Mean Error (m) & 3.72 & 3.32 & 4.55 \\ 
        Std Error (m) & 1.19 & 1.70 & 2.28 \\ 
        Median Error (m) & 3.30 & 2.85 & 4.05 \\ 
        Max Error (m) & 12.00 & 16.15 & 24.65 \\ 
        Min Error (m) & 2.62 & 0.97 & 0.83 \\ 
        \bottomrule
    \end{tabular}
    \label{tab:localization_metrics}
\end{table}

\section{Conclusion and Future work}
\label{conclusion}

This paper presented a cooperative localization framework based on the Unscented Transform (UT) and Split Covariance Intersection (SCI), tailored for scenarios with only single inter-agent range measurements. The UT was leveraged as a technique to account for all sources of uncertainty in generating a state estimate based on the shared information, enabling consistent state estimation despite the limited observability. The SCI method facilitated conservative fusion of shared estimates, ensuring robustness against unknown correlations. Comprehensive Monte Carlo simulations in a 2D environment assessed the impact of various factors on localization accuracy, including range noise, positioning noise, initialization offsets, and motion trajectories. 

Results showed that, while the system maintains localization errors of around 3–4 meters on average, its sensitivity varies across different influencing parameters. In particular, the positioning noise exerted the most dominant influence, with a mean localization error increasing from 2.57 to 10.94 meters. Range noise impacts performance more moderately, with mean error rising from 3.65 to 5.91 meters. The framework also shows resilience to initial position and velocity offsets. Even with increasing offsets, the localization error remained relatively stable, demonstrating robustness in scenarios with imperfect initialization. Notably, degradation in accuracy was gradual and only significant at the highest offset levels. Additionally, path type played a role in performance, with smoother, more continuous paths (e.g., circles and donuts) yielding better accuracy than sharp-edged paths, such as rectangles. Overall, the method exhibits strong performance and stability across a wide range of perturbations, making it suited for deployment in GPS-denied or communication-constrained environments. 

Future work will extend this framework to three-dimensional environments, integrate real-world robotic platforms, and incorporate complementary sensing modalities such as inertial or visual odometry. Additional enhancements will be explored, including adaptive noise modeling and dynamic confidence weighting for SCI, to improve consistency and accuracy under operational constraints.

\section*{Acknowledgments}

This work was supported in part by a research project with Kinnami Software Corporation.

\bibliographystyle{IEEEtran}
\bibliography{references}
\vfill
\end{document}